\documentclass[runningheads]{llncs}
\usepackage[
  left=1in,     
  right=1in,    
  top=1in,      
  bottom=1in    
]{geometry}

\usepackage[T1]{fontenc}
\usepackage{graphicx}

\usepackage{times}
\usepackage{soul}
\usepackage{url}
\usepackage[hidelinks]{hyperref}
\usepackage{amsmath}

\usepackage{booktabs}
\usepackage{algorithm}
\usepackage{algorithmic}
\usepackage{amssymb}
\usepackage{subcaption}
\usepackage{lineno}
\usepackage[numbers]{natbib}
\usepackage{ulem,enumitem} 
\usepackage{subcaption}
\usepackage{caption}
\usepackage{adjustbox}



\usepackage[colorinlistoftodos]{todonotes}

\begin{document}
\title{Conformal Predictive Distributions for Order Fulfillment Time Forecasting}%

\author{Tinghan Ye,
Amira Hijazi, and 
Pascal Van Hentenryck}
\authorrunning{Ye et al.}
%
\institute{H. Milton Stewart School of Industrial \& Systems Engineering, Georgia Institute of Technology, USA
\email{\{joe.ye,ahijazi6,pvh\}@gatech.edu}}
\maketitle              

\begin{abstract}
Accurate estimation of order fulfillment time is critical for e-commerce logistics, yet traditional rule-based approaches often fail to capture the inherent uncertainties in delivery operations. This paper introduces a novel framework for distributional forecasting of order fulfillment time, leveraging Conformal Predictive Systems and Cross Venn-Abers Predictors---model-agnostic techniques that provide rigorous coverage or validity guarantees. The proposed machine learning methods integrate granular spatiotemporal features, capturing fulfillment location and carrier performance dynamics to enhance predictive accuracy. Additionally, a cost-sensitive decision rule is developed to convert probabilistic forecasts into reliable point predictions. Experimental evaluation on a large-scale industrial dataset demonstrates that the proposed methods generate competitive distributional forecasts, while machine learning-based point predictions significantly outperform the existing rule-based system---achieving up to $14\%$ higher prediction accuracy and up to $75\%$ improvement in identifying late deliveries.
\end{abstract}

\section{Introduction}
In the e-commerce industry, rule-based approaches are widely used to estimate delivery dates for ship-to-home order fulfillment. These approaches typically rely on static transit tables that specify carrier transit times, scheduled pickup windows, and valid days for pickup, transit, and delivery. While straightforward to implement, such static configurations often fall short in capturing the dynamic and stochastic nature of real-world logistics operations, particularly as supply chain conditions evolve over time \citep{cheng2025spot}.

\begin{figure*}[!t]
    \centering \includegraphics[width=\textwidth]{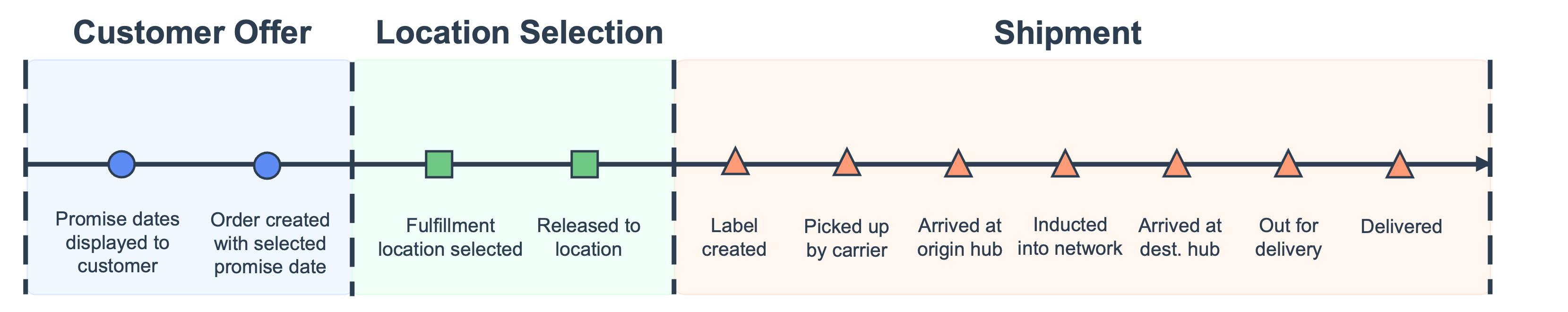}
    \caption{Example of Order Fulfillment Life Cycle.}
    \label{fig:life-cycle}
\end{figure*}

Recently, data-driven approaches--particularly machine learning (ML) methods---have gained traction for estimating delivery times, particularly among large e-commerce companies such as Amazon and JD.com \citep{merchan20242021, ruan2022service}. Despite this increasing adoption, building accurate predictive models for order fulfillment time remains inherently difficult due to the complexity and variability of the fulfillment life cycle. As illustrated in Fig. \ref{fig:life-cycle}, the process begins with the customer offer stage. Once an order is placed, it enters the location selection phase, during which a Fulfillment Location (FL)---such as a warehouse or a retail store---is selected based on inventory availability, customer proximity, and operational constraints. At this stage, a tentative carrier is identified to estimate shipping costs and delivery times, although the final carrier may be confirmed later in the process. The requested stock-keeping unit (SKU) is then retrieved, packed, and staged for pickup according to the operational procedures of the selected fulfillment location. The shipment stage begins with final carrier selection and the generation of a shipping label, marking the point at which the designated carrier assumes responsibility for the order. Last-mile delivery methods can vary, but commonly, the parcel is picked up from the fulfillment location, transported to an origin hub, and moved through the carrier’s network until it reaches the customer.

There is a growing need in the field to move beyond deterministic point predictions and embrace {\it distributional forecasting}, which better captures the uncertainty inherent in order fulfillment time, particularly given the involvement of multiple stakeholders such as retailers, warehouses, and carriers. By modeling the full range of possible outcomes, distributional forecasting offers a more faithful representation of uncertainty and serves as a critical input for contextual stochastic optimization models \citep{ye2024contextual}. 

Beyond logistics, distributional forecasting has gained significant attention in domains such as energy systems \citep{jonkers2024novel, ZHANG2025125369, moradi2025enhanced} and healthcare \citep{arora2023probabilistic}. Compared to quantile or interval forecasts, which represent limited aspects of uncertainty, distributional forecasts provide a complete view of the predictive distribution. This allows for the consistent derivation of point estimates, quantiles, and prediction intervals---capabilities that cannot be recovered in reverse from quantile-based approaches alone \citep{vovk2020computationally}. By more thoroughly quantifying uncertainty, distributional forecasting empowers businesses to adopt risk-aware operational strategies that strike a balance between service reliability and cost efficiency, grounded in rigorous statistical guarantees.

\subsection{Contributions}

This paper introduces a novel framework for the distributional forecasting of order fulfillment time, leveraging conformal prediction, a model-agnostic, distribution-free framework that provides robust coverage/validity guarantees under minimal assumptions. Specifically, Conformal Predictive Systems (CPS) \citep{vovk2020computationally} and Cross Venn-Abers Predictors (CVAP) are applied for probability density forecasting. To the best of the authors' knowledge, this is the first application of CPS and CVAP in delivery service and travel time estimation, establishing a new paradigm for generating reliable predictive distributions. The main contributions are summarized as follows:
\begin{enumerate}
    \item The paper pioneers the use of CPS and CVAP for distributional forecasting in service time estimation, allowing flexible integration with any regression or multi-class classification model, including advanced deep learning architectures.
    \item The paper develops a novel two-stage classify-then-regress model that synergistically combines CPS and CVAP to produce more informative probabilistic forecasts.
    \item The paper proposes a cost-sensitive decision rule to convert conformal-calibrated distributions into flexible point predictions, balancing predictive accuracy with business cost considerations.
    \item The paper introduces granular spatial-temporal features to capture the operational performance of fulfillment locations and carriers, enhancing both predictive power and interpretability.
    \item The paper validates the proposed framework on a real-world industrial dataset comprising more than 5 million orders across over one year, demonstrating that the proposed distributional forecasts are competitive with established methods, while the ML-based point predictions substantially outperform the existing rule-based system.
\end{enumerate}

\noindent
The methodologies proposed in this paper are already being developed and tested by the industrial partner alongside their existing infrastructure. In the initial rollout, they will augment the current fulfillment decision pipeline by providing probabilistic delivery-time guidance in tandem with rule-based estimates.

\subsection{Related Work}
Delivery service time forecasting is a critical aspect of logistics operations. A recent survey on this topic is provided in \citep{wen2024survey}. The specific focus of this paper is parcel delivery service operations in the context of online retail logistics. Prior studies have addressed similar challenges, including \citep{wu2019deepeta, de2021end,  li2021unsupervised, ruan2022service, salari2022real, zhang2023delivery, cao2024revenue,  yi2024learning, zhang2024estimating, ye2024contextual}. While most papers focus on estimating delivery dates at the customer offer stage to set effective customer promise dates, this paper targets predictions at the time of location selection---that is, immediately after order creation but before assignment to a fulfillment location or carrier. This distinction aligns the study more closely with \citep{ye2024contextual}, where estimated delivery dates are leveraged for downstream contextual stochastic optimization to determine a fulfillment location-carrier pair. This work extends beyond prior research by \cite{ye2024contextual} in three significant ways: it introduces novel methodologies for distributional forecasting, incorporates more granular spatio-temporal features particularly related to carrier operations, and validates these advances on a more comprehensive and longitudinal dataset.

Methodologically, this work is closely related to research on probabilistic forecasting for service or travel time estimation. Existing approaches can be generally categorized into parametric and non-parametric methods. Parametric methods assume that uncertainty follows a specific probability distribution \citep{shang2017exploiting, liu2023uncertainty, raj2024stochastic} (e.g., a log-normal distribution as in \citep{liu2023uncertainty}), whereas non-parametric methods make no such assumptions and often employ ML–based approaches. Common ML techniques include tree-based ensemble methods \citep{guo2022forecasting, salari2022real, arora2023probabilistic, ye2024contextual} and deep neural networks \citep{liu2023uncertainty, faulkner2025framework}. These models are typically framed as either Regression or Multi-Class Classification (MC-CLF) problems. Regression-based approaches often rely on specialized model structures, such as Quantile Random Forest (QRF) \citep{meinshausen2006quantile}, to generate predictive distributions \citep{salari2022real}, while classification-based methods naturally produce discrete probability distributions \citep{liu2023uncertainty, ye2024contextual}.

Despite its potential, conformal prediction \citep{vovk2005algorithmic, manokhin_valery_2022_6467205, li2025conformal} has not been widely explored for probabilistic parcel delivery time forecasting. A concurrent study \citep{faulkner2025framework} applies conformal prediction within a classify-then-regress framework and utilizes conformalized quantile regression to construct prediction intervals on a related problem. Other data-driven methods have addressed related operational problems---such as inbound load plan adjustments---by applying two-stage conformal prediction to produce confidence-aware outputs \citep{bruys2024confidence}. However, these methods are limited to prediction set or interval forecasting. In contrast, the framework proposed in this paper explicitly focuses on probability density forecasting, which provides richer information and greater utility for decision-making applications.

This paper also contributes to the literature on cost-centric decision rules, building upon methods previously established by \cite{salari2022real, cao2024revenue} in the context of delivery time estimation.

\section{Problem Definition}

This section revisits the problem of order fulfillment time distributional forecasting, first introduced in \cite{ye2024contextual}. The goal is to estimate the probabilistic distribution of fulfillment time deviation from the customer's promised date, leveraging relevant contextual information. Given a predicted deviation, one can readily compute the estimated delivery date by adding the deviation to the promised date.

Each instance $i$ represents an incoming customer order along with a candidate fulfillment location and an eligible carrier to fulfill and ship the order. Let $\mathcal{X}$ be the covariate (feature) space and $x_i \in \mathcal{X}$ be the covariates associated with instance $i$. Each covariate includes features observable from the order, the fulfillment location, and the carrier. 

Let $\mathcal{Y} = \{c_1, c_2, \ldots, c_K\} \subset \mathbb{Z}$ represent the set of possible fulfillment time deviations, measured in days. The elements of $\mathcal{Y}$ are ordered such that $c_1 < c_2 < \ldots < c_K$, and reflect discrete deviation values relative to the promised delivery date. For each order instance $i$, let $Y_i$ be a random variable that captures the deviation from the promised delivery date, with its realized value denoted by $y_i \in \mathcal{Y}$. Negative values of $y_i$ correspond to early deliveries, zero indicates on-time delivery, and positive values indicate late deliveries.

Given a set of unseen instances $\mathcal{I}$ and their associated covariates, the goal of a forecasting model is to predict the target fulfillment time deviation $y_i$ by producing an estimate $\hat{y}_i$ for each instance $i \in \mathcal{I}$. While traditional approaches focus on point estimation, \textit{distributional forecasting} aims at predicting the full conditional cumulative distribution function (CDF) of the target variable, defined as
\(
F(y_i \mid x_i) = \mathbb{P}(Y_i \leq y_i \mid x_i).
\)
The objective is to learn an estimated function
\(
\hat{F}(y_i \mid x_i) = \hat{\mathbb{P}}(Y_i \leq y_i \mid x_i),
\)
that closely approximates the true conditional CDF. By modeling the entire distribution of possible fulfillment time deviations, distributional forecasting enables more informative and risk-aware decision-making compared to single-point predictions.

Let $\Theta$ denote the set of learnable model parameters, and let $\hat{F}(x_i; \Theta)$ represent the predicted CDF for instance $i$, parameterized by $\Theta$. Let $\mathcal{L}$ be a suitable loss function that measures the discrepancy between the predicted distribution $\hat{F}(x_i; \Theta)$ and the observed outcome $y_i$. Let $\mathcal{D}_{\mathrm{tr}}$ be the set of training instances. The learning task can be formulated as an empirical risk minimization problem:
\[
\min_{\Theta} \ \frac{1}{|\mathcal{D}_{\mathrm{tr}}|} \sum_{i \in \mathcal{D}_{\mathrm{tr}}} \mathcal{L}\left(\hat{F}(x_i; \Theta), y_i \right).
\]

One key advantage of distributional forecasting is that it enables efficient and flexible uncertainty quantification without requiring model retraining or recalibration. Specifically, any desired quantile of the predicted fulfillment time deviation can be extracted directly from the predicted CDF, supporting a wide range of operational decisions.
The $q$-th quantile ($q \in [0, 100]$) is defined as:
\(
\hat{y}_i^q = \inf\left\{y: \hat{\mathbb{P}}(Y_i \leq y \mid x_i) \geq q \right\}.
\)

This formulation allows practitioners to tailor predictions based on risk preferences and service-level objectives. For instance, a lower quantile (e.g., $q = 10$) can be used to generate \textit{conservative estimates} of delivery time, useful for setting aggressive internal targets or evaluating best-case scenarios.  The median ($q = 50$) provides a \textit{typical expected deviation}, often used for customer-facing delivery promises.  An upper quantile (e.g., $q = 90$) offers a \textit{pessimistic but cautious estimate}, helpful for planning buffer stock or adjusting carrier dispatch windows.

Moreover, prediction intervals can be constructed by selecting appropriate quantile pairs. For example, a 95\% prediction interval for the fulfillment time deviation is given by:
\(
[\hat{y}_i^{2.5}, \hat{y}_i^{97.5}],
\)
which captures the range within which the deviation is expected to fall with high confidence. In a logistics context, such intervals are valuable for \textit{capacity planning}, \textit{carrier selection}, and \textit{customer communication}, as they provide a data-driven way to quantify delivery risk under uncertainty.

\section{Data Description}

This study utilizes an anonymized e-commerce transactional order dataset provided by an industrial partner. The dataset contains over 5 million representative ship-to-home orders collected between July 2023 and November 2024. For demonstration purposes, the analysis focuses on orders shipped via a heavily utilized ground shipping services recommended by the industrial partner. However, the proposed framework is designed to be readily extensible to other carriers and service types. To mitigate the influence of extreme outliers, orders with fulfillment time deviations beyond the range of $[-10, 10]$ days—that is, delivered more than 10 days earlier or later than expected—were excluded from the dataset.

The dataset's features can be categorized into three groups: order-level, fulfillment location-level, and carrier-level features. In addition to the features outlined in \cite{ye2024contextual}, this study incorporates additional data for each category, including customer membership tier and various timestamps within the fulfillment location and carrier network, making the dataset significantly more comprehensive. Following feature engineering (detailed in Section \ref{sec:feat-eng}), all features utilized in the ML models were summarized in Table \ref{tab:all_features} in the Appendix.

Fig. \ref{fig:time-dist} illustrates the empirical distribution of fulfillment time deviations for orders shipped using the carrier examined in this study. The distribution is centered around zero, indicating that a substantial proportion of orders are delivered on time. However, a noticeable spread exists on both sides, reflecting instances of early and late deliveries. In logistics operations, any deviation---whether ahead of or behind the promised delivery date---can disrupt planning, reduce customer satisfaction, and introduce inefficiencies into downstream processes \cite{salari2022real, cui2024sooner}.

\begin{figure}[!t]
\centering
    \centering
    \includegraphics[width=0.6\textwidth]{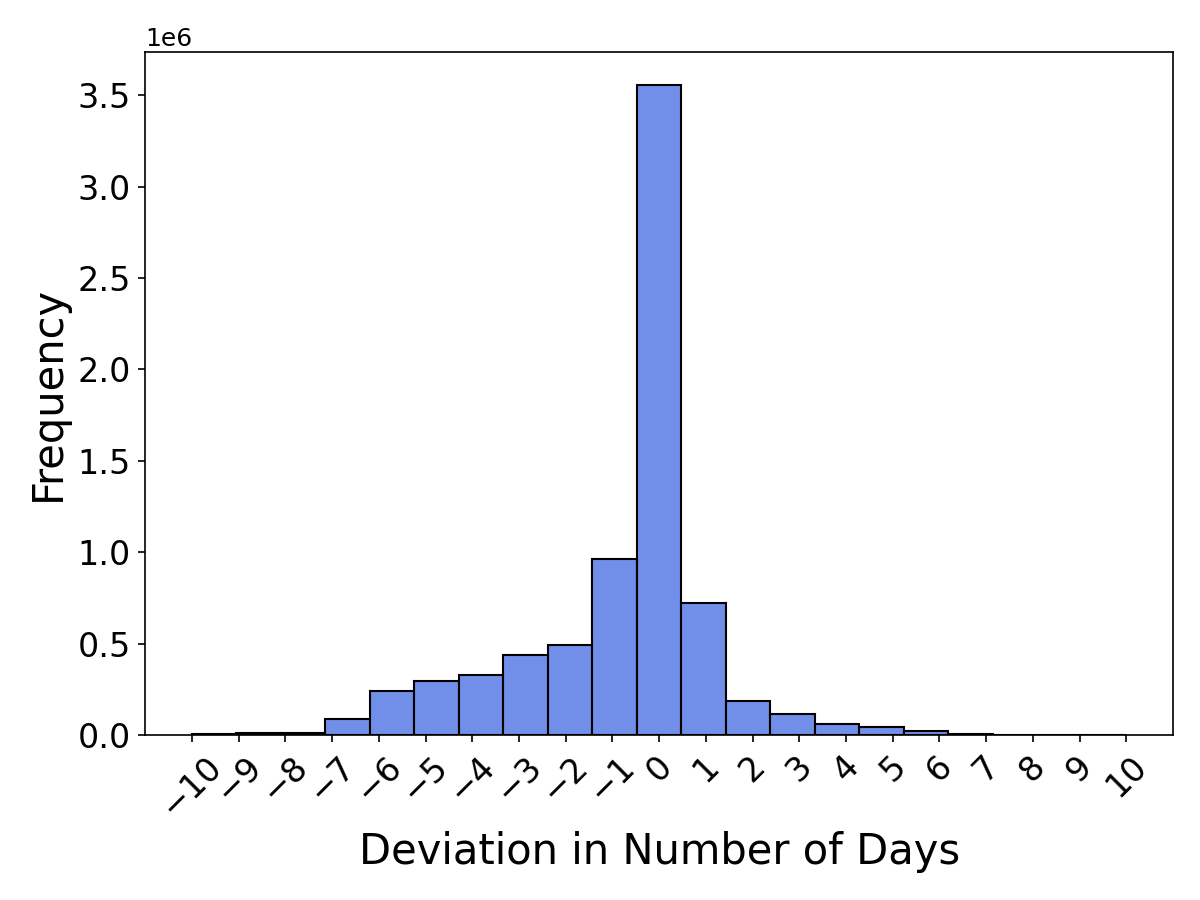}
    \caption{Distribution of Fulfillment Time Deviations for Orders Shipped via the Selected Carrier.}
    \label{fig:time-dist}
\end{figure}

\section{Methodology}
\begin{figure*}[!t]
    \centering
\includegraphics[width=\textwidth]{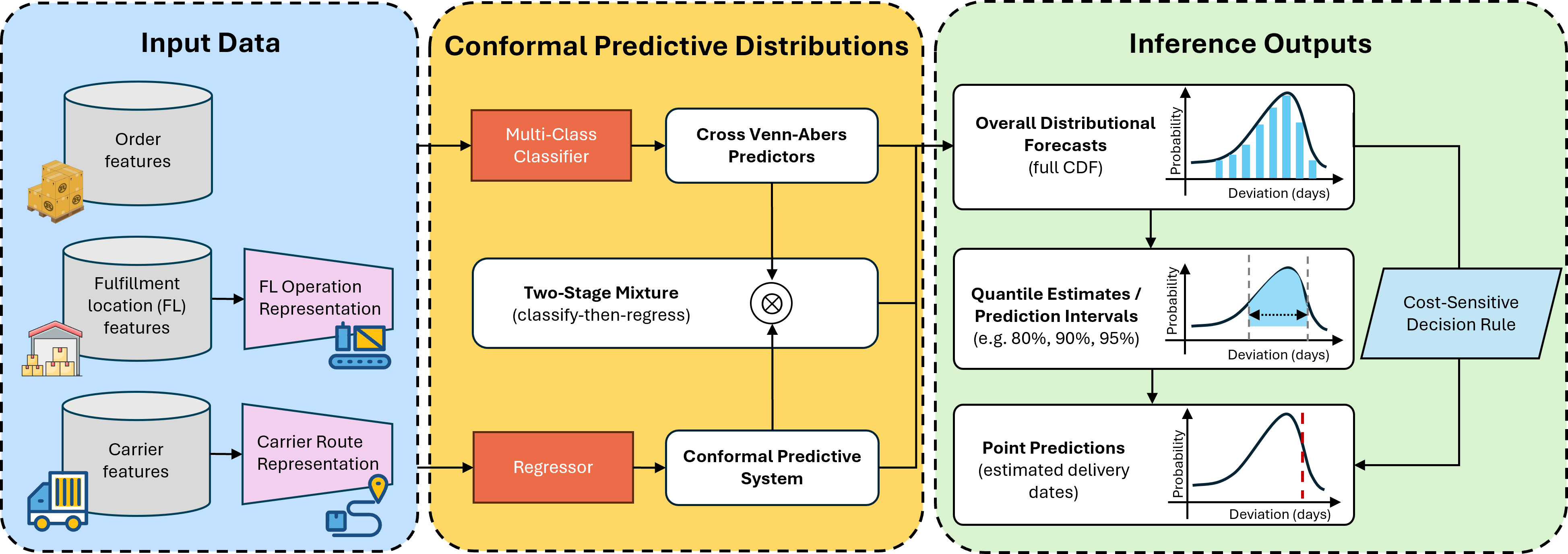}
    \caption{An Overview of the Conformal Predictive Distribution Framework for Order Fulfillment Time Forecasting.}
    \label{fig:framework}
\end{figure*}

This section introduces the proposed forecasting framework, illustrated in Fig. \ref{fig:framework}. The framework first constructs dynamic spatio-temporal representations from raw features, which serve as inputs to either a deterministic regression model or a multi-class classification (MC-CLF) model. To quantify uncertainty, conformal prediction techniques are applied to calibrate the underlying models, generating distributional forecasts---with Conformal Predictive System (CPS) used for regression models and Cross Venn-Abers Predictors (CVAP) for MC-CLF models. Additionally, the framework incorporates an optional cost-sensitive decision rule, which converts conformal calibrated probability distributions into point predictions, ensuring alignment with operational objectives.

\subsection{Dynamic Spatio-Temporal Representation} \label{sec:feat-eng}
The first step of the implementation transforms raw inputs into representative features for ML models. These features capture information about the fulfillment locations and the carrier routes. 

\subsubsection{Fulfillment Location Operation Representation}
To represent real-time congestion in each location, the ML models include a feature that captures the waiting time at the time of order placement. To quantify workload intensity, the shipping volume in the past two hours is extracted as a proxy for recent operational throughput. For seasonal workload patterns, the ML models leverage the domain expertise from industrial practitioners to segment each FL’s operation time based on wave schedules, operational hours, and facility type. Within each segment, summary statistics of historical processing times were computed using training data. 

\subsubsection{Carrier Route Representation}

Since the carrier plays a crucial role in the shipment stage of order fulfillment, the proposed framework integrates a spatio-temporal representation of carrier performance, derived from anonymized historical carrier tracking data. This representation captures key routing characteristics by aggregating information on shipment patterns, carrier hub activity, and transit times.
 For each origin-destination (O-D) pair, defined at the ZIP5-to-ZIP5 level, the framework extracts aggregated metrics to estimate typical carrier routes. Shipment metrics include the total number of shipments, median number of intermediate stops, and median number of tracking updates per shipment. Exception metrics quantify disruptions, such as the number of shipments delayed by the carrier, weather, or other issues.

To model hub dynamics, the most frequently used origin and destination hubs are identified for each O-D pair. Their latitude and longitude are recorded, and the Haversine distance between them is calculated to represent the common spatial path. Temporal characteristics are captured through the average and standard deviation of key time intervals, including the time from label creation to the initial scan at the origin hub, the induction time into the carrier network, the transit time from the origin scan to the destination scan, and the times from destination scan to ``out for delivery” and from ``out for delivery” to final delivery. For unseen O-D pairs in the test set, the framework uses a hierarchical imputation strategy, aggregating geographic units from ZIP5 to ZIP3, and then to ZIP2 levels, ensuring robust route estimation based on the most granular available data.

\subsection{Conformal Predictive Distribution for Regression and Classification} \label{sec:conformal}

\subsubsection{Conformal Predictive System} \label{sec:cps}

The \textit{Conformal Predictive System for Regression} (CPS) is a flexible, model-agnostic framework for constructing predictive distributions that offer formal uncertainty quantification. Unlike traditional approaches that rely on strong distributional assumptions or asymptotic guarantees, CPS provides finite-sample, distribution-free coverage under the mild assumption of exchangeability(i.e., the order of data points doesn't affect the underlying probability distribution)~\cite{vovk2020computationally}. This makes it particularly well-suited for real-world applications—such as logistics—where the data-generating process is often complex, non-stationary, or only partially understood.

In logistics operations, decisions are frequently made under uncertainty, whether one is forecasting fulfillment times, estimating shipment delays, or allocating resources across stochastic demand. CPS enables such decisions to be informed, not only by point predictions, but also by valid, data-driven confidence estimates that reflect the model's empirical error behavior. By transforming regression outputs into full predictive distributions, CPS supports robust risk assessment and reliability-aware planning.

The following paragraphs describe two implementations of this framework: the \textit{Split Conformal Predictive System (SCPS)}, which offers globally calibrated uncertainty estimates, and the \textit{Mondrian Conformal Predictive System (MCPS)}, which introduces local adaptation to improve calibration in heterogeneous prediction settings.

\paragraph{Split Conformal Predictive System}

The SCPS is a model-agnostic framework that transforms any regression model’s point estimate into a full predictive distribution with finite-sample coverage guarantees. This is particularly useful in logistics settings, where outcome variability—such as delays or fulfillment deviations—can arise from unpredictable operational conditions.

\begin{enumerate}
    \item \textbf{Data Splitting:}
    Partition the training dataset \( \mathcal{D}_{\text{tr}} = \{(x_i, y_i)\}_{i=1}^{N}\) into two disjoint subsets: a \textit{proper training set} \( \mathcal{D}_t \), used to train the regression model; a \textit{calibration set} \( \mathcal{D}_c = \{(x_j, y_j)\}_{j=1}^{N_c} \), used to quantify prediction uncertainty. This separation allows the model to be evaluated on unseen data when constructing its prediction intervals, improving generalization.

    \item \textbf{Model Training:}
    Train a regression model \( h: \mathcal{X} \rightarrow \mathbb{R} \) using the proper training set \( \mathcal{D}_t \). The output \( h(x) \) is a point prediction for a new input \( x \), i.e., the expected fulfillment time deviation.

    \item \textbf{Calibration Score Computation:}
    For a new input \( x \) and for each calibration example \( (x_j, y_j) \in \mathcal{D}_c \), compute the calibration score
    \(
    C_j = h(x) + (y_j - h(x_j)),
    \)
    where \( y_j - h(x_j) \) represents the prediction error for a historical instance. By adding this residual to the prediction for the test instance \( h(x) \), the method generates a set of plausible outcomes that reflect how the model has erred in the past. This technique aligns with practical logistics intuition: it adjusts today’s prediction based on past deviations, under the assumption that similar mistakes might reoccur.

    \item \textbf{Predictive Distribution Construction:}
    Sort the calibration scores to obtain an ordered set:
    \(
    C_{(1)} < C_{(2)} < \dots < C_{(N_c)},
    \)
    and define boundaries \( C_{(0)} = -\infty \), \( C_{(N_c+1)} = \infty \). Let \( \tau \sim \text{Uniform}(0,1) \) be a random variable used to smooth the predictive distribution. Then, the estimated CDF can be computed as follows:
    \[
    \hat{F}(y \mid x) =
    \begin{cases}
        \frac{n+\tau}{N_c+1} & \text{if } y \in (C_{(n)}, C_{(n+1)})\text{ for } n \in \{0, \ldots N_{c}\}, \\
        \frac{n'-1 + (n'' - n' + 2)\tau}{N_c+1} & \text{if } y = C_{(n)} \text{ for } n \in \{1, \ldots N_{c}\},
    \end{cases}
    \] where    \(
    n' = \min\{m : C_{(m)} = C_{(n)}\} \text{ and }
    n'' = \max\{m : C_{(m)} = C_{(n)}\}.
    \)

    This function \( \hat{F}(y \mid x) \) represents the probability that the true outcome will be less than or equal to \( y \), based on the empirical distribution of historical model errors. As a result, it provides calibrated, data-driven uncertainty estimates. In logistics contexts, this allows for the construction of predictive intervals that quantify risks in terms of fulfillment deviations. For instance, a narrow interval centered around zero indicates high model confidence that a shipment will arrive on time, while a wider interval may suggest increased likelihood of delay—prompting preemptive interventions such as route adjustments or customer notifications.
\end{enumerate}

\paragraph{Mondrian Conformal Predictive System}

While SCPS provides a globally calibrated predictive distribution, it assumes that the model’s prediction errors are uniformly distributed across all inputs. In many real-world settings, and particularly in logistics, this assumption may not hold. For example, the uncertainty associated with predicting short, on-time deliveries may differ markedly from that of predicting extreme delays. The MCPS addresses this by partitioning the calibration data according to the model’s own predictions, thereby producing \textit{locally adaptive} predictive distributions \citep{bostrom2021mondrian}. MCPS introduces the following modification to the SCPS framework:

\begin{enumerate}
    \item \textbf{Partitioning by Predicted Output:}
    Define \( B \in \mathbb{N} \) as the number of bins (a tunable hyperparameter). Partition the calibration set \( \mathcal{D}_c = \{(x_j, y_j)\}_{j=1}^{N_c} \) into \( B \) disjoint subsets \( \mathcal{D}_{c}^1, \ldots, \mathcal{D}_{c}^B \), based on the predicted values \( h(x_j) \). Each bin \( \kappa \in \{1,\ldots,B\}\)  corresponds to a range of predicted deviations. For example, one bin may cover near-zero (on-time) predictions, while others may correspond to moderate or extreme delays. This allows MCPS to tailor uncertainty quantification to the model’s confidence in different prediction regimes.

    \item \textbf{Calibration Score Computation (Local):}
    For a test input \( x \), determine the bin \( \kappa \) to which \( h(x) \) belongs. Then compute calibration scores using only the data in the corresponding subset  \( \mathcal{D}^{\kappa}_{c} = \{(x_{j}, y_{j})\}_{j=1}^{N^{\kappa}_{c}} \), as follows:
    \(
    C^{\kappa}_{j} = h(x) + (y^{\kappa}_{j} - h(x^{\kappa}_{j})) \text{ for } j = \{1, \dots, N_{c}^{\kappa}\}.
    \)
    As in SCPS, these scores simulate plausible outcomes for the test case by adjusting its prediction using the historical errors observed in similar prediction contexts.

    \item \textbf{Predictive Distribution Construction (Local):}
Sort the bin-specific calibration scores:
    \(
    C_{(1)}^{\kappa} < C_{(2)}^{\kappa}  < \dots < C_{(N_{c}^{\kappa})}^{\kappa} ,
    \)
    and define boundary values \( C_{(0)}^{\kappa} = -\infty \), \( C_{(N_{c}^{\kappa}+1)}^{\kappa} = \infty \). Let \( \tau \sim \text{Uniform}(0,1) \) as before.
Then, the bin-specific CDF is:
    \[
    \hat{F}(y \mid x) =
    \begin{cases}
        \frac{n+\tau}{N_{c}^{\kappa}+1} & \text{if } y \in (C_{(n)}^{\kappa}, C_{(n+1)}^{\kappa}) \text{ for } n \in \{0, \ldots N_{c}^{\kappa}\}, \\
        \frac{n'-1 + (n'' - n' + 2)\tau}{N_{c}^{\kappa}+1} & \text{if } y = C_{(n)}^{\kappa} \text{ for } n \in \{1, \ldots N_{c}^{\kappa}\},
    \end{cases}
    \]
where    \(
    n' = \min\{m : C_{(m)}^{\kappa} = C_{(n)}^{\kappa}\} \text{ and }
    n'' = \max\{m : C_{(m)}^{\kappa} = C_{(n)}^{\kappa}\}.
    \)

    This localized predictive distribution reflects the empirical error behavior for predictions in the same range as the test input. In contrast to SCPS, which pools all calibration scores, MCPS only uses scores from cases where the model made similar predictions. In logistics, this enables more refined uncertainty quantification. For instance, the model may be highly confident in predicting on-time deliveries, resulting in narrower intervals for such cases. Conversely, predictions associated with rare or extreme delays may yield wider intervals, signaling greater uncertainty. This differentiation can support risk-sensitive operational decisions, such as prioritizing high-risk shipments for additional monitoring or contingency planning.
\end{enumerate}

\subsubsection{Venn-Abers Calibration for Probabilistic Multi-Class Classification (MC-CLF)} \label{sec:va}

This section addresses probabilistic multi-class classification (MC-CLF), where the target variable takes values from a finite discrete set \( \{c_1, \dots, c_K\} \). Predicting a conditional CDF in this setting is equivalent to estimating a categorical probability distribution:
\(
p_v = \mathbb{P}(Y = c_v \mid x), ~\text{for } v \in \{1, \dots, K\}.
\) Raw outputs from many ML models often do not represent well-calibrated probabilities, necessitating transformation via calibration methods.

\paragraph{Isotonic Regression Calibration}

 Isotonic Regression (IR) \citep{zadrozny2002transforming, ye2024contextual} is a non-parametric method which is a common baseline approach for MC-CLF. For binary classification ($K=2$), IR uses a scoring function \( s: \mathcal{X} \rightarrow [0, 1]\), which represents the uncalibrated probability estimate for the positive class.
Given $s$, IR learns a non-decreasing step-function $f_{\text{IR}}: [0,1] \rightarrow [0, 1]$ that maps a raw model score $s(x)$ to a calibrated probability $\hat{\mathbb{P}}(Y=1 \mid x)$. This function is fitted to minimize error (e.g., squared loss) on a calibration dataset $\mathcal{D}_c = \{(x_j, y_j)\}_{j=1}^{N_c}$. Essentially, IR sorts instances by increasing scores, and finds the best-fitting monotonically increasing piecewise constant function (staircase). While useful, standard IR lacks finite-sample theoretical guarantees. 

\paragraph{Inductive Venn-Abers Predictor (IVAP)}

As a theoretically robust alternative, \textit{Venn-Abers Predictors} (VAP) leverage ideas from IR and the conformal prediction framework~\cite{vovk2015large, manokhin2017multi}. VAPs offer distribution-free, finite-sample calibration guarantees under the mild assumption of exchangeability. The Inductive VAP (IVAP) modifies the standard IR approach to produce a probability \textit{interval}, explicitly quantifying uncertainty. The IVAP procedure for binary classification (K = 2) is specified as follows:

\begin{enumerate}
    \item \textbf{Data Splitting:} Divide the training dataset \( \mathcal{D}_{\text{tr}} \) into two subsets: a proper training set \( \mathcal{D}_t \) and a calibration set \( \mathcal{D}_c = \{(x_j, y_j)\}_{j=1}^{N_c} \).
    \item \textbf{Model Training:} Train a binary classification model on \( \mathcal{D}_t \), yielding a scoring function \( s: \mathcal{X} \rightarrow [0, 1]\), where $s(x)$ represents the uncalibrated probability estimate for the positive class.
    \item \textbf{Dual Isotonic Regressions:} For a new instance $x$ with score $s(x)$, construct the interval by fitting two separate IR models: $f_0$ and $f_1$. \( f_0 \), is fitted on \( \{(s(x_j), y_j)\}_{j=1}^{N_c} \cup \{(s(x), 0)\} \), and \( f_1 \), is fitted on \( \{(s(x_j), y_j)\}_{j=1}^{N_c} \cup \{(s(x), 1)\} \). These define the prediction interval
\[
(p_0, p_1) := \left(f_0(s(x)), f_1(s(x))\right),
\]
which brackets the possible calibrated probabilities based on the two possible labels for the test instance. This dual fitting process prevents overly confident probability estimates. If the model is confident, \( p_0 \) and \( p_1 \) will be close together. A wider interval indicates greater uncertainty in the model output.

\end{enumerate}

\paragraph{Cross-Validated Venn-Abers Predictor (CVAP)}
To enhance robustness over a single data split and obtain a single point estimate, the CVAP applies the IVAP procedure within a $k$-fold cross-validation scheme. This yields $k$ probability intervals $[p_0^{(f)}, p_1^{(f)}]$ for a test instance. These intervals are then aggregated into a single probability estimate using a minimax-style geometric mean rule:
\begin{equation} \label{eq:GM}
    p = \frac{\text{GM}(p_1)}{\text{GM}(1 - p_0) + \text{GM}(p_1)},
\end{equation}
where \( \text{GM}(p_1) = \left(\prod_{f=1}^k p_1^{(f)}\right)^{1/k} \) and similarly for \( \text{GM}(1 - p_0) \). This aggregation balances information across all intervals and avoids overconfidence by integrating lower and upper bounds from each fold. The result is a single, calibrated probability estimate with strong empirical and theoretical support.

\paragraph{Extension to Multi-Class Classification}

Venn-Abers calibration is extended to multi-class problems ($K > 2)$ using a one-versus-rest strategy. This involves training $K$ distinct binary classifiers, where the $i$-th classifier distinguishes class $i$ from all other classes. These binary classifiers are then individually calibrated with CVAP, yielding $K$ probability estimates $p^1, \dots, p^K$. As these are generated independently, a final normalization step ensures they form a valid probability distribution:
\(
\hat{p}_v = p^v / \sum_{v'=1}^{K} p^{v'},
\)
for each class $v \in \{1, \ldots, K\}$.

\subsection{A Two-Stage Method: Classify-then-Regress with Truncated Conformal Prediction}

This section proposes a two-stage forecasting method for modeling fulfillment time deviations that combines \textit{Venn-Abers-calibrated classification} with \textit{conformal predictive system for regression}. The method is specifically designed to handle the mixed nature of the deviation variable: many orders arrive exactly on time, while the rest exhibit continuous variability in early or late arrivals. The proposed method—referred to as the \textit{truncation-based classify-then-regress approach}—generates a full probability distribution over the delivery deviation for each order.  It effectively integrates a categorical structure by first classifying orders into three distinct groups: early, on-time, or late. Within each of these categories, it then captures the continuous variability by predicting the specific magnitude of deviation. To ensure reliable probabilistic estimates, the approach incorporates conformal regression and Venn-Abers predictors. This combination enables both interpretable classification and precise, well-calibrated regression predictions.

\paragraph{Step 1: Classification Stage — Predicting Delivery Status}

Define a discrete random variable \( S_i \in \{-1, 0, 1\} \) representing the delivery status:
\[
S_i = \begin{cases}
    -1 & \text{if } Y_i < 0 \quad \text{(early)}, \\
     0 & \text{if } Y_i = 0 \quad \text{(on-time)}, \\
     1 & \text{if } Y_i > 0 \quad \text{(late)}.
\end{cases}
\]
A multi-class classifier is trained and calibrated using the Venn-Abers procedure, returning the class probabilities
\(
P_c(S = s \mid x), ~ \text{for each } s \in \{-1, 0, 1\}.
\)

\paragraph{Step 2: Regression Stage — Predicting Deviation Magnitude with Truncation}

A conformal regression model (SCPS or MCPS, see Section~\ref{sec:cps}) is trained on the full dataset. For each calibration example \( (x_j, y_j) \), define the calibration score for a test input \( x \) with model prediction \( h(x) \) as
\(
C_j = h(x) + (y_j - h(x_j)).
\)
These scores simulate realistic outcomes by combining model predictions with observed residuals. Let the sorted scores be
\(
C_{(1)} < C_{(2)} < \dots < C_{(N_c)},
\)
and identify the index \( m \) and $n$ such that $m = \max\{m: C_{(m)} < 0 \}$ and $n = \min \{n: C_{(n)} > 0\} $. At prediction time, the following truncation strategy is applied to generate class-consistent predictive distributions:
\begin{itemize}
    \item \textbf{On-time deliveries} (\( S = 0 \)): Define \( P_r(y \mid x, S = 0) \) as a degenerate distribution concentrated at 0. That is, the model assigns all probability mass to the outcome \( y = 0 \).
    
    \item \textbf{Early deliveries} (\( S = -1 \)): Construct the predictive distribution \( P_r(y \mid x, S = -1) \) using only scores \( \{C_{(1)}, \dots, C_{(m)}\} \), i.e., scores corresponding to negative deviations.
    
    \item \textbf{Late deliveries} (\( S = 1 \)): Construct the distribution \( P_r(y \mid x, S = 1) \) with scores \( \{C_{(n)}, \dots, C_{(N_c)}\} \), i.e., scores greater than zero.
\end{itemize}
This truncation ensures consistency between the predicted status and the continuous deviation forecast, without discarding training data as in class-specific regression models.

\paragraph{Step 3: Mixture Formation — Integrating Classification and Regression}

Before combining the components, recall that each class-conditional distribution \( P_r(y \mid x, S = s) \) captures how far off a delivery might be, assuming the shipment falls into class \( s \). The classifier returns a probability for each class. To get the full predictive distribution, average the class-conditional distributions using the class probabilities as weights:
\[
\hat{F}(y \mid x) = \sum_{s \in \{-1, 0, 1\}} P_c(S = s \mid x) \cdot P_r(y \mid x, S = s).
\]

\paragraph{Alternative Strategy and Justification}

A straightforward alternative would be to train separate regression models and apply CPS calibration within each class. However, this strategy underperforms in practice due to \textit{selection bias}: classification errors cause mismatches between training and prediction regimes, especially near class boundaries. The truncation-based approach addresses the limitations of separate regression models by using the full dataset during training and calibration, rather than dividing data based on predicted classes. This avoids the selection bias introduced by classification errors, which often lead to mismatches between training and prediction conditions near class boundaries. Instead of applying class-specific constraints during training, the truncation-based method enforces these constraints only at inference time. This strategy enhances robustness and ensures that calibration guarantees are preserved across the full range of predictions.

\subsection{Obtaining Point Predictions from Distributional Forecasts}

While the proposed framework focuses on generating full predictive distributions for fulfillment time deviations, many logistics systems continue to operate with single-point estimates for decision-making tasks. Moreover, point predictions are necessary for comparing the performance of distributional models against legacy systems that do not support uncertainty quantification. This section outlines several methods for deriving point predictions from distributional forecasts, ranging from standard model-specific outputs to cost-sensitive decision rules tailored to real-world priorities.

\paragraph{Default Point Prediction Methods}

For CPS-based methods, a straightforward and principled choice is the median (50\% quantile) of the predicted CDF. This corresponds to the point that minimizes expected absolute error. For MC-CLF models, the predicted class with the highest probability is selected:
\(
\hat{y} = \arg\max_j \mathbb{P}(Y = c_j \mid x),
\)
representing the most probable delivery day.

\paragraph{Cost-Sensitive Decision Rule}

To go beyond fixed quantiles, this paper introduces a cost-sensitive decision rule for selecting a quantile that minimizes a weighted loss function reflecting practical trade-offs. The procedure is:
\begin{enumerate}
    \item Fit CPS and/or CVAP models to obtain predictive distributions.
    \item On a validation set, perform a grid search over \( \eta \in [0,100] \).
    \item For each candidate \( \eta \), extract point predictions \( \hat{y}^\eta \) as the \( \eta \)\%-quantile of the CDF.
    \item Evaluate the following cost-sensitive objective:
    \[
    \min_{\eta \in [0, 100]} \text{RMSE}(\hat{y}^\eta) + \beta~ \text{Late\_RMSE}(\hat{y}^\eta) + \gamma ~\text{Early\_RMSE}(\hat{y}^\eta),
    \]
\end{enumerate}
where \text{RMSE} is the overall root mean squared error,
\text{Late\_RMSE} corresponds to the RMSE over late deliveries, and \text{Early\_RMSE} captures the RMSE for early deliveries. The penalty weights \( \gamma \) and \( \beta \) reflect user-defined preferences for penalizing early or late deliveries. The optimal quantile \( \eta^* \) is then applied to each test instance $i$ to produce final point predictions:
\[
\hat{y}_i = \inf\left\{y: \hat{\mathbb{P}}(Y_i \leq y \mid x_i) \geq \eta^* \right\}.
\]
This formulation enables practitioners to tailor prediction strategies according to operational priorities, such as minimizing customer dissatisfaction due to early or late arrivals.

\section{Experiment Results}
This section provides experimental results on the proposed methods. The evaluation uses  large-scale industrial instances, and the proposed methods are compared with several baseline methods.

\subsection{The Experimental Setting}

\begin{figure}[!t]
    \centering
    \includegraphics[width=0.6\textwidth]{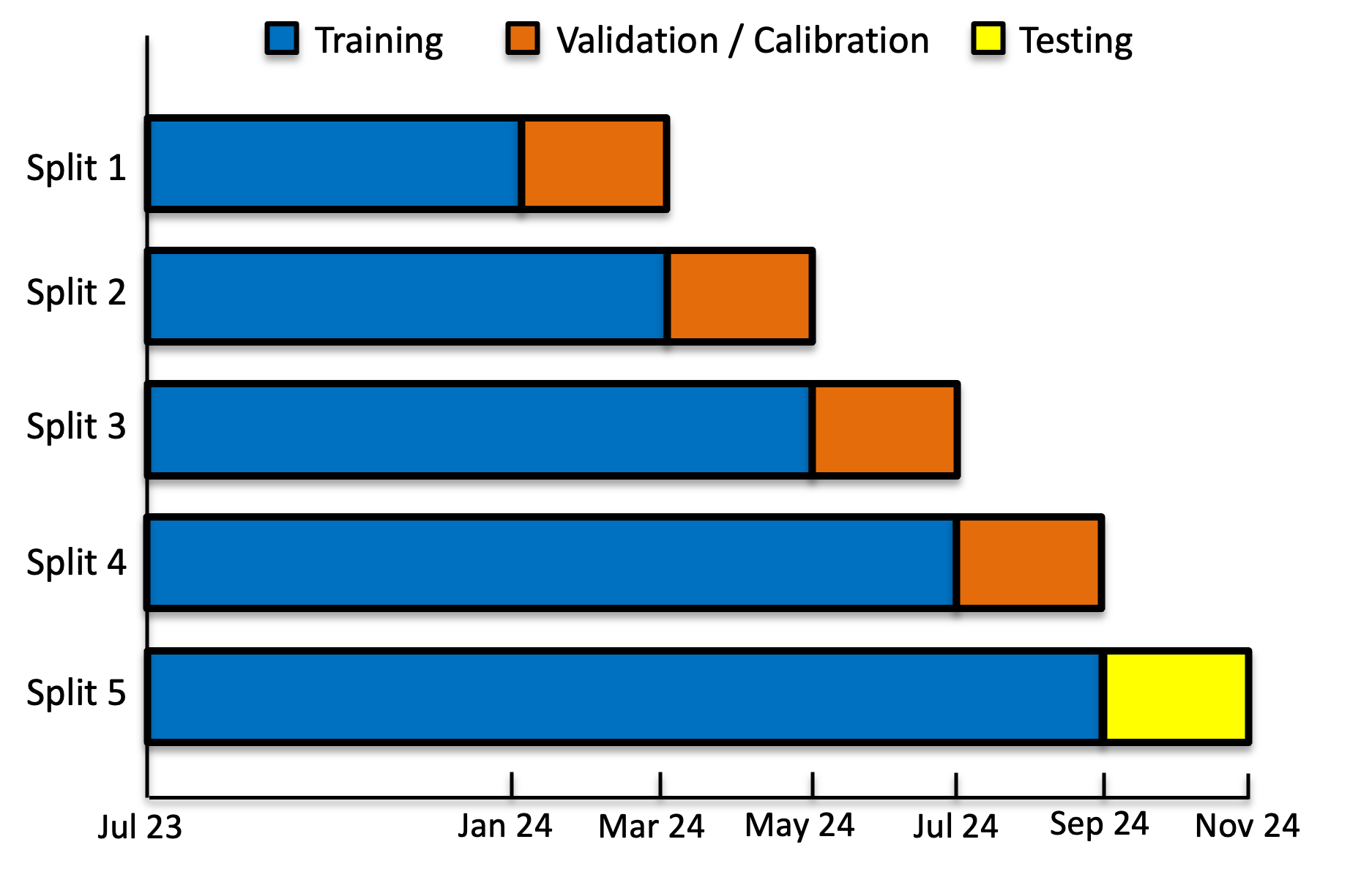}
    \caption{Data Splitting for the Experimental Results.}
    \label{fig:data-split}
\end{figure}

\paragraph{Data Splitting Organization}
The dataset was partitioned using a 4-fold time-series cross-validation approach to preserve temporal dependencies, as illustrated in Fig. \ref{fig:data-split}. The validation sets were consistently used across all training runs to tune the hyperparameters of the ML models. For CVAP, calibration sets from all four folds were utilized, whereas for SCPS and MCPS, only the calibration set from the last fold was used. To ensure a fair evaluation, the last two months of data were reserved for testing. Within the test set, only orders that were shipped via the same carrier service that was estimated at the time of location selection were retained. This restriction ensures a direct and unbiased comparison between the proposed models and the existing system.

\paragraph{Tools and Models}
The CPS-based models were implemented using Crepes \cite{bostrom2022crepes}, while the CVAP was adapted from the Venn-ABERS Calibration library. For the MCPS, the number of bins was set to 10 to balance between model flexibility and computational efficiency. XGBoost \cite{chen2016xgboost} was used as the underlying regression model for all CPS methods, while CatBoost \cite{prokhorenkova2018catboost} was employed for MC-CLF.  Both models are widely adopted in industry and demonstrated superior performance in initial experiments compared to other standard machine learning models, aligning with the findings in \cite{ye2024contextual}. Optuna was used for hyperparameter tuning for all ML models \cite{akiba2019optuna}.

\paragraph{Feature Importance}

\begin{figure}[!t]
\centering
    \centering
    \includegraphics[width=\textwidth]{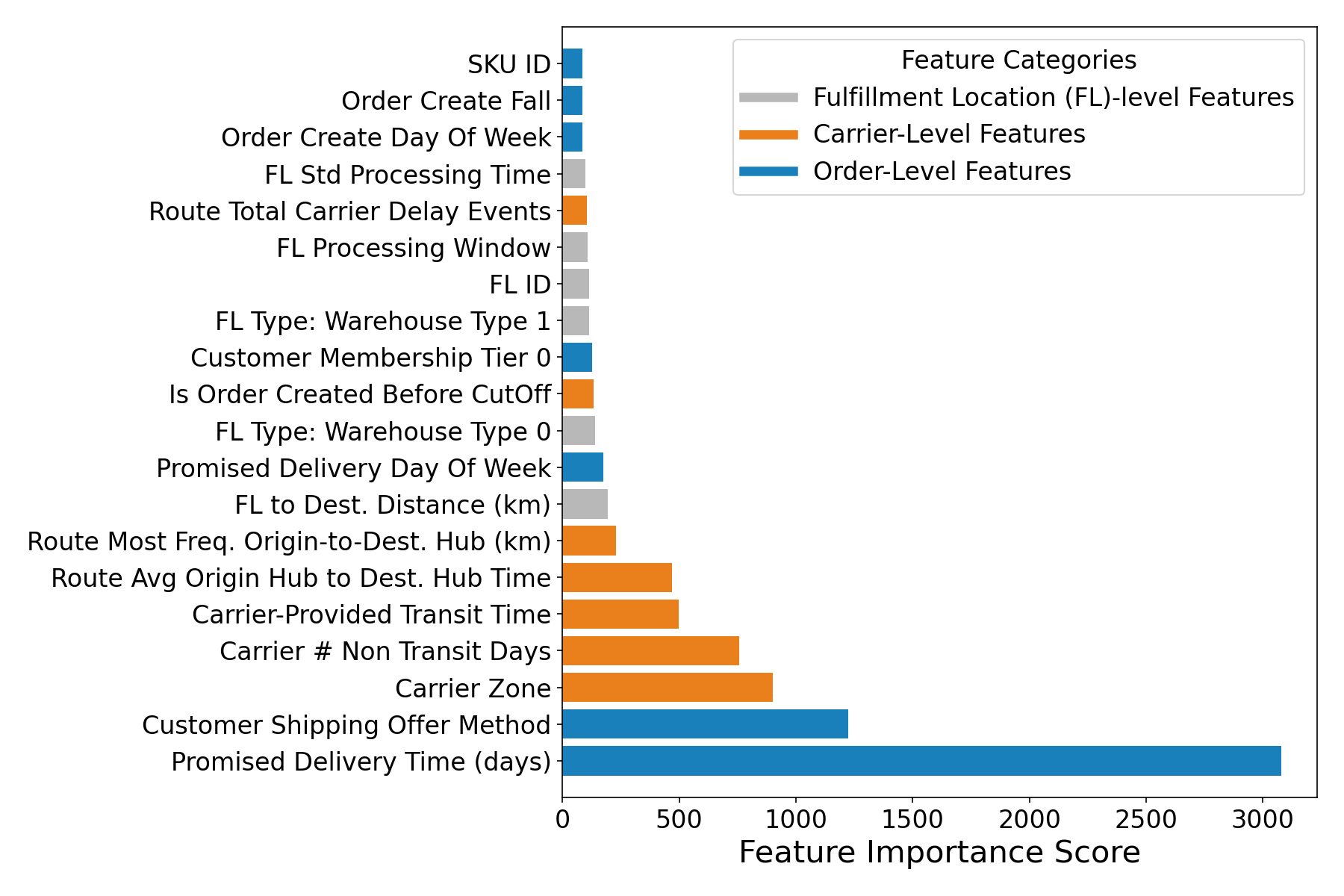}
    \caption{The Top 20 Most Important Features used by the XGBoost Regressor.}
    \label{fig:feat-imp}
\end{figure}

Interpretability plays a crucial role in the deployment of ML models in industrial applications. Tree-based ML models inherently provide feature importance scores by aggregating the gains from feature splits.  Fig. \ref{fig:feat-imp} illustrates the top 20 most influential features identified by the XGBoost regression model. Order-level features rank highest in importance, but route representation and FL operation features also significantly impact predictions. Key route-related features include average transit time between hubs and frequently used hub-to-hub distances, while FL operation features like wave time segment and processing time contribute to fulfillment time variability. These findings confirm the necessity of incorporating both route and FL operation representations in effective fulfillment time forecasting.

\paragraph{Evaluation Metrics}
Table \ref{tab:evaluation_metrics} summarizes the evaluation metrics considered in the experiments. In particular, the continuous ranked probability score (CRPS) measures the accuracy of the probability density forecast. The pinball loss (PL) and the mean absolute quantile coverage error (MQCE) measure the sharpness and reliability of the quantile estimates.  The accuracy, RMSE, and the mean absolute error (MAE) evaluate the point predictions. The accuracy measures the number of times that an order is delivered exactly on the predicted date. RMSE and MAE measure the differences in days between the predicted and the correct dates.

\begin{table*}[!t]
\scriptsize
\centering
\caption{Summary and Description of Evaluation Metrics.}
\begin{adjustbox}{width=\textwidth}
\label{tab:evaluation_metrics}
\begin{tabular}{l|l|l}
\toprule
\textbf{Metric} & \textbf{Formula} & \textbf{Description} \\ \hline
CRPS & $ \frac{1}{N_{test}}\sum_{i=1}^{N_{test}} \int_{-\infty}^{\infty} (\hat{F}(u) - \mathbf{1}\{u \ge y_i\})^2 du$ & Continuous ranked probability score.\\ \hline
PL & $\displaystyle 
\begin{aligned}
\frac{1}{9}\sum_{q \in [0.1,0.2,\ldots,0.9]} &\frac{1}{N_{test}}\sum_{i=1}^{N_{test}} \Bigl( q\, \max\{y_i- \hat{y}^{100q}_i, 0\} \\
&+ (1 - q)\, \max\{\hat{y}^{100q}_i - y_i, 0\} \Bigr)
\end{aligned}$ & Average pinball loss. \\ \hline
MQCE & $\frac{1}{9}\sum_{q \in [0.1, 0.2, \ldots, 0.9]} |\{\frac{1}{N_{test}}\sum_{i=1}^{N_{test}}\mathbf{1}_{y_i < \hat{y}^{100q}_i}\} - q|$  & Mean quantile coverage error. \\ \hline
Accuracy & $ \frac{1}{N_{test}}\sum_{i=1}^{N_{test}} \mathbf{1}(y_i = \hat{y}_i)$ & Number of correct predictions over the total number of predictions. \\ \hline
RMSE & $\sqrt{ \frac{1}{N_{test}}\sum_{i=1}^{N_{test}}(\hat{y}_i - y_i)^2}$ & Root mean squared error (in number of days). \\ \hline
MAE & $ \frac{1}{N_{test}}\sum_{i=1}^{N_{test}}|\hat{y}_i - y_i|$ & Mean absolute error (in number of days). \\ \hline
\end{tabular}
\end{adjustbox}
\end{table*}

\subsection{Distributional Forecasting Performance}
\begin{figure*}[!t]
\centering
    \centering
    \includegraphics[width=\textwidth]{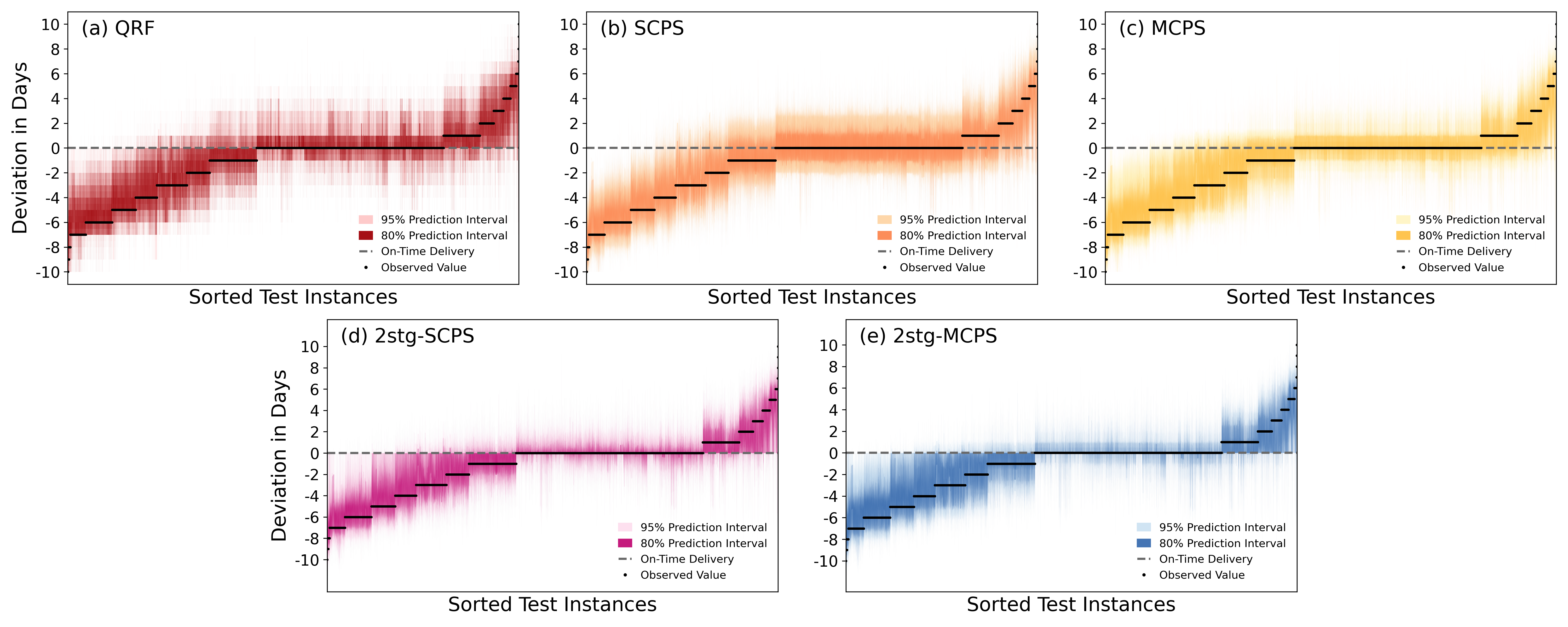}

    \caption{Prediction Intervals on Sorted Test Set Instances.}
    \label{fig:intervals}
\end{figure*}

Table \ref{tab:distribution_metrics} summarizes the distributional forecasting performance on the test set: it reports the PL, MQCE, CRPS, and the coverage and mean interval sizes for three symmetric prediction intervals (80\%, 90\%, 95\%). The implementations of Quantile Random Forest (QRF) and Isotonic Regression (IR)-calibrated MC-CLF closely follow the modeling approaches in \cite{ye2024contextual}. Fig. \ref{fig:intervals} shows the prediction intervals generated by the regression-based and two-stage methods.

Overall, the calibrated MC-CLF models outperform their uncalibrated counterparts in CRPS by around $16\%$. The CVAP calibration shows competitive performance with IR calibration, with marginally better PL and MQCE. Notably CVAP yields higher coverage rates, though neither calibration approach perfectly matches the targeted coverage levels at all intervals. While calibrated classifiers excel in distribution accuracy, CPS-based regression methods demonstrate superior quantile estimate reliability through lower pinball loss ($5\%$ to $10\%$ lower than QRF) and better interval coverages (all reaching nominal coverages). MCPS produces on average $0.1$ to $0.3$ days narrower intervals than SCPS with stronger distributional metrics due to its adaptive binning.

Two-stage methods (2stg-SCPS/MCPS) combine these strengths, achieving remarkable sharpness. 
As shown in Fig. \ref{fig:intervals}(d–e), their 80\%, 90\% and 95\% bands are visibly the tightest of all -- on average $0.2$–$0.4$ days narrower than those from QRF in panel (a) -- yet they do so without “wasting” coverage on the wrong side of zero. For the earliest deliveries (left of the zero-error line), virtually all of the interval mass lies strictly below zero, while for the latest shipments (right of zero) the bands shift are entirely above. The on-time cases densely cluster around zero with very narrow intervals, delivering pinpoint estimates of punctual performance. This targeted contraction translates directly into better scores: up to an $8.7\%$ reduction in CRPS versus single-stage CPS and an $11$–$13\%$ improvement in pinball loss relative to QRF, all while producing the narrowest uncertainty intervals overall. The trade-off is a modest dip in marginal coverage and a slightly looser quantile calibration compared to full CPS; in practice, it dramatically cuts unnecessary buffer days and yields far more actionable forecasts.

\begin{table*}[!t]
\centering
\scriptsize
\caption{Average Pinball Loss, MQCE, CRPS, Coverage, and Interval Sizes on the Test Set.}
\label{tab:distribution_metrics}
\begin{adjustbox}{width=0.7\textwidth}
\begin{tabular}{l|ccc|ccc|ccc}
\toprule
\textbf{Methods}
& \textbf{PL $\downarrow$} & \textbf{MQCE $\downarrow$} & \textbf{CRPS $\downarrow$} 
& \multicolumn{3}{c|}{\textbf{Coverage}} & \multicolumn{3}{c}{\textbf{Mean Interval Size $\downarrow$}} \\
&  &  &  
& 80\% & 90\% & 95\% & 80\% & 90\% & 95\% \\
\midrule
Base MC-CLF & 0.335 & 0.232 & 0.498 & 0.427 & 0.600 & 0.768 & 2.043 & 2.749 & 3.517 \\
IR MC-CLF   & 0.298 & 0.240 & \textbf{0.413} & 0.619 & 0.776 & 0.901 & 2.232 & 2.944 & 3.659 \\
CVAP MC-CLF & 0.297 & 0.232 & \textbf{0.419} & 0.658 & 0.835 & 0.935 & 2.312 & 3.049 & 3.780 \\
\midrule
QRF         & 0.273 & 0.313 & 0.502 & \uline{0.876} & \uline{0.923} & \uline{0.965} & \textbf{1.844} & \textbf{2.587} & \textbf{3.445} \\
\midrule
SCPS        & 0.261 & \textbf{0.035} & 0.482 & \uline{0.817} & \uline{0.907} & \uline{0.953} & 2.133 & 3.374 & 4.582 \\
MCPS        & \textbf{0.244} & \textbf{0.041} & 0.451 & \uline{0.829} & \uline{0.915} & \uline{0.953} & 2.097 & 3.002 & 4.200 \\
\midrule
2stg-SCPS   & \textbf{0.238} & 0.217 & \textbf{0.440} & 0.795 & 0.879 & 0.935 & \textbf{1.439} & \textbf{2.264} & \textbf{3.172} \\
2stg-MCPS   & \textbf{0.242} & \textbf{0.213} & 0.449 & \uline{0.828} & 0.886 & 0.925 & \textbf{1.664} & \textbf{2.391} & \textbf{3.135} \\
\bottomrule
\multicolumn{9}{l}{\scriptsize\textit{Note: $\downarrow$ indicates lower values are better. Best three values are \textbf{bolded}.}}\\
\multicolumn{9}{l}{\scriptsize\textit{Coverage values that achieve nominal rates (80\%, 90\%, 95\%) are \uline{underlined}.}}
\end{tabular}
\end{adjustbox}
\end{table*}

\subsection{Point Prediction Performance}

This section evaluates the point prediction performance of each model. The objective is to produce a point prediction that can be directly compared to the actual fulfillment date. 

\paragraph{Methods for Generating a Single Estimated Delivery Date}
The methods used to derive a single estimated delivery date can be summarized as follows:
\begin{itemize}
    \item {\it Current System}: The point prediction of the existing system which is a rule-based system using static transit time tables and cutoff times.
    
    \item {\it MC-CLF (Base/IR/CVAP)}: The point prediction is given by the MC-CLF model predicting the most likely day-offset.
    
    \item {\it CVAP MC-CLF-Rule}: A MC-CLF model calibrated by CVAP, which uses a cost-sensitive decision rule for quantile selection before rounding to the nearest day.
    
    \item {\it Base Reg}: The point prediction is obtained by a standard regression predicting the mean deviation, rounded to the nearest day.
    
    \item {\it SCPS/MCPS-Median}: The point prediction is the median of the CPS rounded to the nearest day.
    
    \item {\it SCPS/MCPS-Rule}: Applies cost-sensitive decision rule on the CPS for quantile selection before rounding to the nearest day.
    
    \item {\it 2-stg-SCPS/MCPS-Rule}: The point prediction is obtained by applying a cost-sensitive decision rule on the two-stage model for quantile selection before rounding to the nearest day.
\end{itemize}
For all models using a cost-sensitive decision rule, the default parameter values are $\beta = 0.5$ and $\gamma = 0$. These settings prioritize the accurate prediction of late deliveries over early ones, reflecting a common operational objective, i.e., the minimization of late arrivals, which typically incur higher penalties or customer dissatisfaction.
 
\subsubsection{Key Observations}
Table \ref{tab:point_metrics2} reports the point prediction metrics, highlighting the overall accuracy, RMSE, MAE, the percentage of late deliveries identified, and error metrics specifically on late deliveries.

The current system's sub-$50\%$ accuracy reveals significant room for data-driven improvement, particularly in identifying late deliveries. This limitation is unsurprising, given that the test set reflects historical decisions where the chosen fulfillment location was already deemed acceptable. As a result, the system is biased toward on-time predictions and performs poorly under dynamic or uncertain conditions. 

In contrast, ML-based approaches enhance the overall accuracy by up to $14\%$, identify substantially more late deliveries (up to $75\%$), and reduce associated error metrics by up to $0.6$ days overall and up to $1.5$ days on actual late deliveries. Both IR and CVAP improve overall accuracy by around $5\%$ compared to their uncalibrated version. The uncalibrated MC-CLF performs poorly on most metrics relative to other ML methods. By contrast, the calibrated MC-CLF models strike a better balance between accuracy and late-delivery detection (up to $13\%$ higher than Base MC-CLF). CPS methods offer stronger calibration for point predictions than a simple regression baseline. Indeed, comparing SCPS-/MCPS-Median with Base Reg reveals gains in overall accuracy ($2\%$ to $3\%$ higher) and error metrics ($0.01$ to $0.03$ days lower). However, their performance on strictly late deliveries can be less clear-cut. Since calibration primarily refines the typical (mean or median) estimation, it may not always help with the tail behavior tied to late arrivals. 

\begin{figure}[!t]
    \centering
     \begin{subfigure}[!t]{0.45\textwidth}
        \centering
\includegraphics[width=\textwidth]{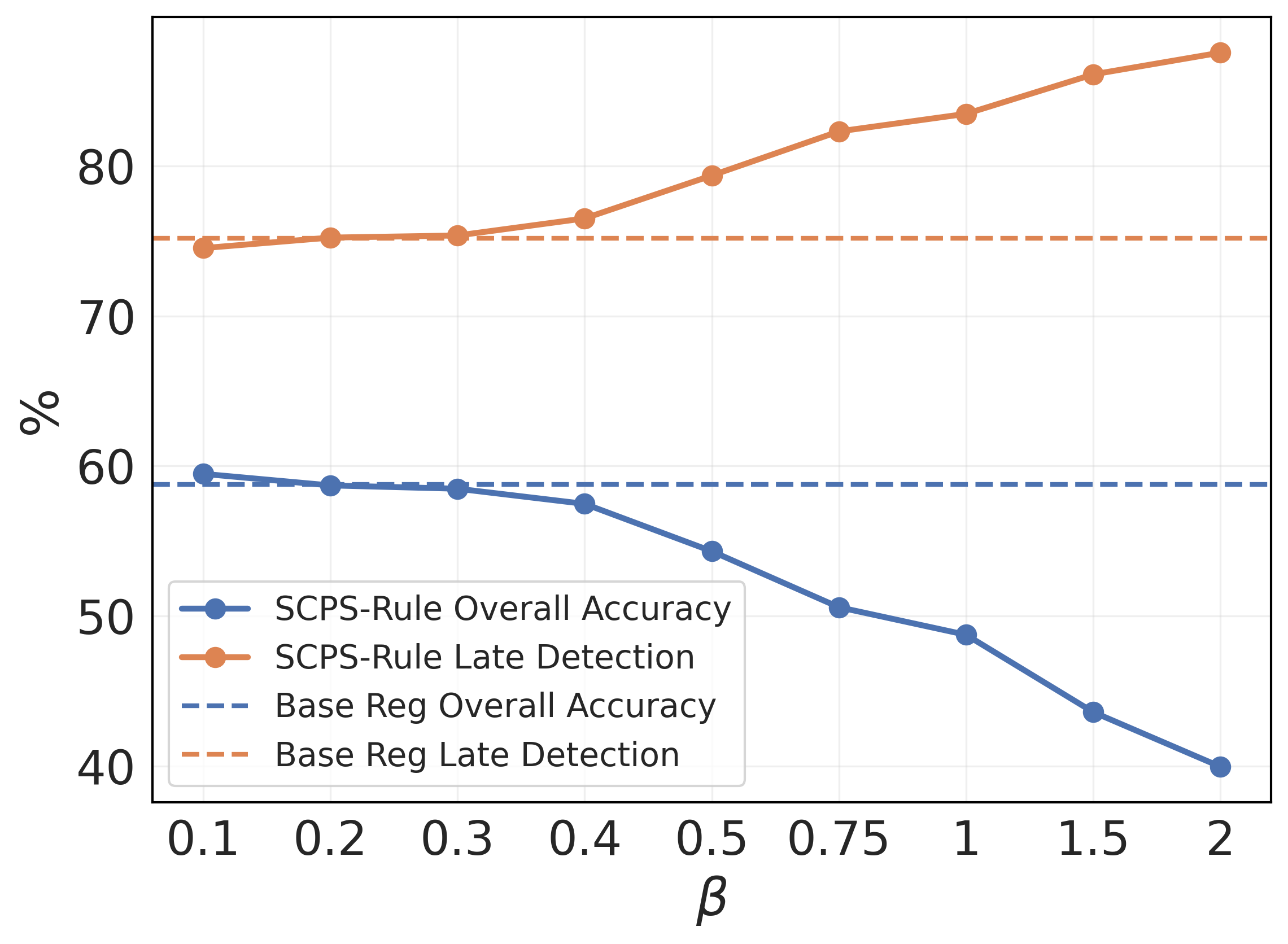}
    \end{subfigure}
    \begin{subfigure}[!t]{0.45\textwidth}
        \centering
\includegraphics[width=\textwidth]{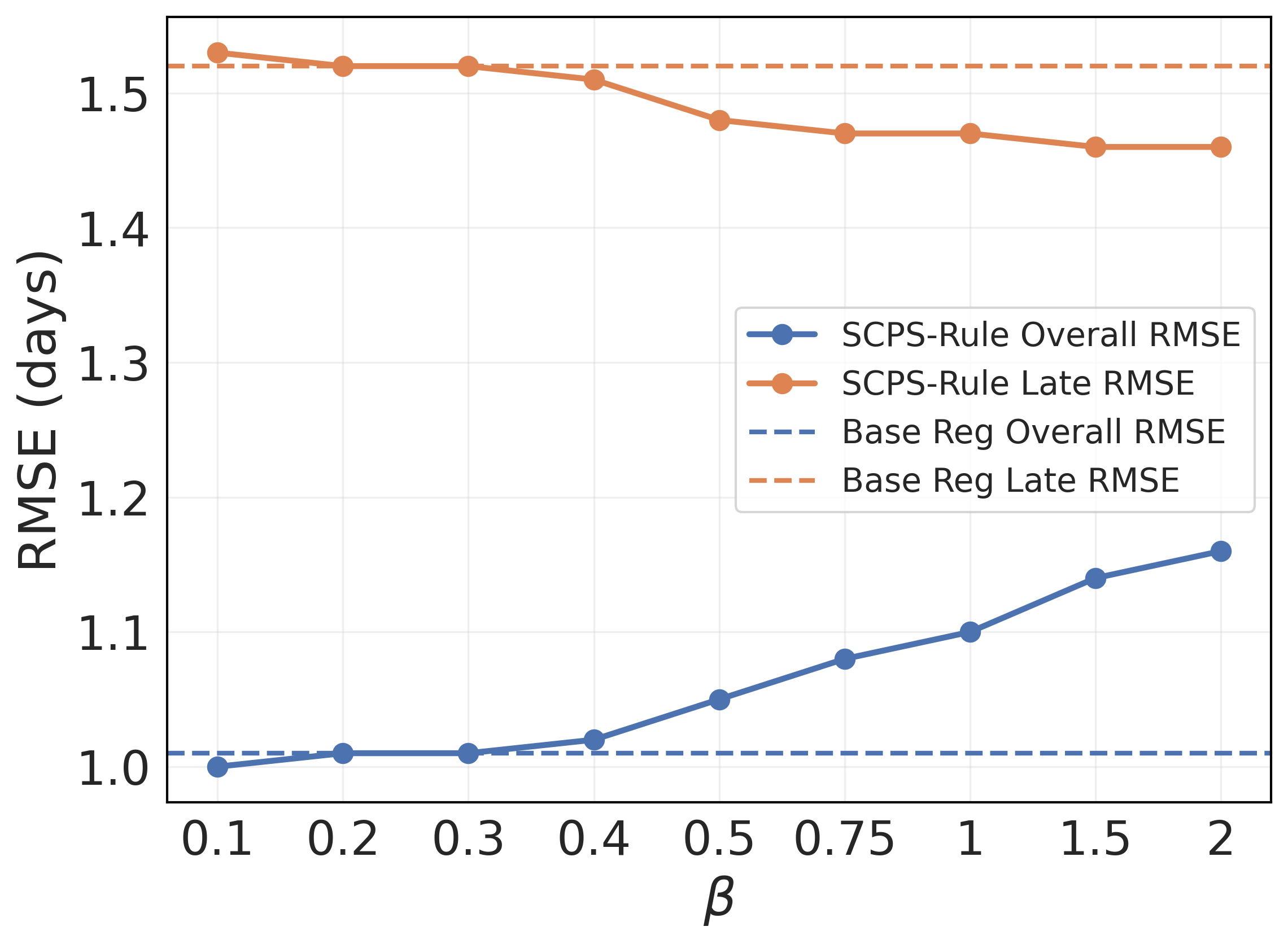}
    \end{subfigure}    
    \caption{Accuracy / RMSE vs. Late Delivery Weight ($\beta$) in the Decision Rule Objective.}
    \label{fig:sensitivity_beta}
\end{figure}

Introducing a cost-sensitive decision rule---particularly SCPS- and MCPS-Rule---shifts the emphasis toward more aggressively predicting late shipments. This leads to higher late-detection rates ($1.5\%$ to $4.2\%$ higher than Base Reg) and lower late-delivery errors (up to $0.05$ days lower than Base Reg), albeit at the cost of overall accuracy ($4\%$ to $8\%$ lower than Base Reg). As shown in Fig. \ref{fig:sensitivity_beta}, the degree of this trade-off can be fine-tuned by adjusting the weight $\beta$ in the decision rule’s objective function, allowing the system to balance penalties for late misclassification against the desire for a high overall prediction accuracy. 
When applied to two-stage methods however, these decision rules demonstrate reduced effectiveness. This diminished performance likely stems from the two-stage methods' inherent prioritization of tighter prediction intervals over quantile estimation reliability.

\begin{table*}[!t]
\scriptsize
\caption{Point Prediction Metrics on the Test Set.}
\centering
\begin{adjustbox}{width=\textwidth}

\begin{tabular}{lcccccc}
\toprule
\textbf{Methods} & \textbf{Overall Accuracy(\%) $\uparrow$
} & \textbf{Overall RMSE} $\downarrow$ & \textbf{Overall MAE} $\downarrow$ & \textbf{Late Detection(\%)} $\uparrow$  &  \textbf{Late RMSE} $\downarrow$ & \textbf{Late MAE} $\downarrow$\\  
\midrule
Current System & 49.53 & 1.66 & 1.01 & 3.82 & 2.99 & 2.49
\\ 
\midrule
Base MC-CLF & 58.50 & 1.30 & 0.71 & 48.67 &	1.84 & 1.31
\\
IR MC-CLF & \textbf{64.41}  & 1.14 & 0.58 & 63.60 & 1.71 & 1.09\\
CVAP MC-CLF	& \textbf{64.14}		&1.16	&0.59	&62.53		&1.72	&1.11\\
CVAP MC-CLF-Rule	&52.47	&1.30	&0.77	&73.60	&1.62	&1.09\\
\midrule
Base Reg & 58.79 &  \textbf{1.01} & \textbf{0.56} & \textbf{75.17} & \textbf{1.52} & \textbf{0.97}
\\
SCPS-Median &60.23 & \textbf{1.00}	& \textbf{0.54}	&73.37		&1.54	&0.98\\
MCPS-Median & \textbf{61.47}	& \textbf{1.00}	& \textbf{0.53}	&63.21	&1.60	&1.07\\
SCPS-Rule ($\beta = 0.5$) & 54.32	& 	1.05	&0.62	& \textbf{79.37}	& \textbf{1.48}	& \textbf{0.95}\\
MCPS-Rule ($\beta = 0.5$)	&	50.63	&1.12	&0.68	& \textbf{76.66}		& \textbf{1.47}	& \textbf{0.93}\\
\hline
2stg-SCPS-Rule & 58.83 & 1.08 & 0.6 & 66.71 & 1.56 & 1.04\\
2stg-MCPS-Rule & 58.18 & 1.08 & 0.6 & 59.99 & 1.59 & 1.08\\
\bottomrule
\multicolumn{7}{l}{
\footnotesize\textit{ Note: Arrows indicate whether higher ($\uparrow$) or lower ($\downarrow$) values are better.} }\\ 
\multicolumn{7}{l}{
\footnotesize\textit{ Bold numbers indicate top three values for each metric. } } 
\end{tabular}
\label{tab:point_metrics2}
\end{adjustbox}
\end{table*}

\section{Discussion and Managerial Implications}

The value of distributional forecasting extends beyond statistical accuracy to the quality of the downstream decisions it enables. By quantifying the full spectrum of potential delivery outcomes, the proposed framework allows managers to proactively manage risk. The optimal forecasting model, however, is contingent on the specific managerial objective.

\subsubsection{Informing Stochastic and Robust Optimization} For risk-neutral stochastic optimization, where the goal is to optimize expected outcomes, CVAP MC-CLF is highly valuable. It produces calibrated distributional forecasts with validity guarantees, which can inform decisions such as selecting a carrier or fulfillment location \cite{ye2024contextual}. For risk-averse or robust optimization, SCPS/MCPS methods are more suitable, generating prediction intervals with strong coverage guarantees. These intervals can construct the uncertainty sets over which a decision is robustified. This is crucial for applications like establishing reliable delivery windows that serve as core constraints for vehicle routing, or guiding strategic network decisions, such as identifying high-performance warehouses to ensure sufficient inventory allocation.

\subsubsection{Integrating with Operational Workflows} For direct integration into rule-based operational workflows, the choice of model depends on system capabilities. Two-stage models produce more actionable intervals; by first predicting a delivery status (e.g., late), they then generate a deviation interval conditional on that outcome, which facilitates transparent triage rules. For legacy systems requiring point forecasts, SCPS/MCPS-Median offers strong risk-neutral accuracy. For cost-sensitive decisions where late deliveries must be minimized, SCPS/MCPS-Rule provides targeted control, with a tunable $\beta$ weight to flexibly manage the trade-off. This is directly applicable to proactive customer communication, where a conservative forecast can trigger early notifications, or staffing decisions, where anticipated delays can justify scheduling contingent labor.

\section{Conclusions}
This study presents a novel framework for distributional forecasting of delivery times, leveraging conformal prediction techniques to provide robust and interpretable uncertainty estimates. By incorporating granular spatio-temporal features, the proposed approach enables more accurate and adaptable probabilistic predictions. The introduction of a cost-sensitive decision rule further ensures that point predictions align with business objectives, balancing accuracy and the detection of late deliveries. 

Importantly, the industrial partner has expressed strong interest in deploying something very similar to the proposed system. Components of this system are under active development to be tested against the current system. Initially, the proposed system will most likely be used to augment the existing location-selection process by providing delivery-time guidance alongside the current rule-based estimation infrastructure.

For future work, a key direction is integrating this framework with omnichannel fulfillment optimization (e.g., \cite{ye2024contextual}) to evaluate their impact on downstream tasks, such as location and carrier selection. Fulfillment time forecasting with coverage or validity guarantees could provide provable bounds on the objective of the downstream optimization problem, enhancing decision-making reliability. In addition, there are several promising ways to extend the forecasting models. One approach is to integrate deep learning techniques with conformal predictive distributions. Another is to enrich the input space by incorporating additional features, such as autoregressive time-series components \citep{tang2022applying} or weather-related information \citep{ZHANG2025125369}.

\paragraph{\ackname} 
This research was partly supported by the NSF AI Institute for Advances in Optimization (Award 2112533)

\bibliographystyle{splncs04}
\bibliography{refs}

\newpage
\section*{Appendix}
\begin{table*}[!ht]
\scriptsize
\centering
\caption{Features Used in the Machine Learning Model across Order, Fulfillment Location, and Carrier Levels.}
\label{tab:all_features}
\begin{adjustbox}{width=\textwidth}
\begin{tabular}{l | l | l}
\toprule
\textbf{Feature Name} & \textbf{Description} & \textbf{Type} \\ 
\midrule
\multicolumn{3}{c}{\textbf{Order-Level Features}} \\
\midrule
Order Create Month/Day/Hour & Date-time components when order was created & Numerical \\
Order Create Season/Holiday & Season and holiday indicator of order date & Categorical/Binary \\
Promised Delivery Month/Day & Date-time components of promised delivery & Categorical \\
Promised Delivery Holiday & Whether promised date is a holiday & Binary \\
Promised Delivery Time & Days from order to promised delivery & Numerical \\
Shipping Offer Method & Fulfillment experience at order time & Categorical \\
Order Total Units, Weight, Dim. Weight & Size and quantity metrics of the order & Numerical \\
SKU ID/Units/Height/Length/Width/Weight & SKU attributes and quantity in order & Mixed \\
SKU Dept/Class/Subclass/Perf. Code & SKU category and performance level & Categorical \\
Customer Membership Tier & Customer membership type & Categorical \\
Dest. ZIP/State & Customer location & Categorical \\
\midrule
\multicolumn{3}{c}{\textbf{Fulfillment Location (FL)-Level Features}} \\
\midrule
FL ID/Type/Lat/Lon & Fulfillment location identifier and position & Categorical/Numerical \\
FL to Dest. Distance & Distance to customer & Numerical \\
SKU On-Hand Inventory & SKU availability at location & Numerical \\
FL Capacity Status & Whether FL is over capacity & Binary \\
FL Waiting Units / Same SKU & Queue length for all and same SKU & Numerical \\
FL Shipped Units Last 2Hrs / Same SKU & Recent shipping activity & Numerical \\
FL Processing Window & Time window of order creation & Categorical \\
FL Avg/Std Processing Time & Processing time stats & Numerical \\
\midrule
\multicolumn{3}{c}{\textbf{Carrier-Level Features}} \\
\midrule
Transit Time / Zone / Cutoff Time & Carrier time and region details & Numerical \\
\# Non Transit Days / Before Cutoff & Service schedule and timing compliance & Mixed \\
Route Use Freq. / Median \# Stops/Updates & Route-level shipment frequency & Numerical \\
Total Carrier/Weather/Other Exceptions & Exception event counts & Numerical \\
Avg/Std: Label to Hub, Hub Induction, Hub Transit & Timings between key shipping stages & Numerical \\
Avg/Std: Dest. Scan to Delivery, Final Delivery & Delivery segment durations & Numerical \\
Most Freq. Origin/Dest. Hub & Common route endpoints & Categorical \\
Most Freq. Hub Lat/Lon, Distance & Geolocation and distance between hubs & Numerical \\
\bottomrule
\end{tabular}
\end{adjustbox}
\end{table*}

\end{document}